\documentclass[letterpaper, 10 pt, conference]{ieeeconf}  % Comment this line out if you need a4paper

\IEEEoverridecommandlockouts                              % This command is only needed if 
\usepackage{xcolor}

\usepackage{graphicx}              % For including images
\usepackage{subcaption}            % For subfigures
\usepackage{url}
\usepackage{comment}

\title{\LARGE \bf
Continuous Cardiac Arrest Prediction in ICU \\ using PPG Foundation Model
}

\author{Saurabh Kataria$^{1}$, Ran Xiao$^{1}$, Timothy Ruchti$^{2}$, Matthew Clark$^{2}$, Jiaying Lu$^{1}$, \\ Randall J. Lee$^{3}$, Jocelyn Grunwell$^{4}$, and Xiao Hu$^{1}$% <-this % stops a space
\thanks{$^{1}$ Nell Hodgson Woodruff School of Nursing, Emory University.
        Correspondence: {\tt\small saurabh.kataria@emory.edu}}%
\thanks{$^{2}$ Nihon Kohden Digital Health Solutions, Inc
        {\tt\small }}%
 \thanks{$^{3}$ University of California, San Francisco{\tt\small }}%
  \thanks{$^{4}$ Children's Healthcare of Atlanta{\tt\small }}%
}

\begin{document}
\maketitle
\thispagestyle{empty}   % normal to have this
\pagestyle{empty}

%%%%%%%%%%%%%%%%%%%%%%%%%%%%%%%%%%%%%%%%%%%%%%%%%%%%%%%%%%%%%%%%%%%%
% mention why this work is imp. for clinicians
\begin{abstract}
Non-invasive patient monitoring for tracking and predicting adverse acute health events is an emerging area of research.
We pursue in-hospital cardiac arrest (IHCA) prediction using \emph{only} single-channel finger photoplethysmography (PPG) signals.
Our proposed two-stage model Feature Extractor-Aggregator Network (FEAN) leverages powerful representations from pre-trained PPG foundation models (PPG-GPT of size up to 1 Billion) stacked with sequential classification models.
We propose two FEAN variants (``1H'', ``FH'') which use the latest one-hour and (max) 24-hour history to make decisions respectively.
Our study is the first to present IHCA prediction results in ICU patients using only unimodal (continuous PPG signal) waveform deep representations.
With our best model, we obtain an average of 0.79 AUROC over 24~h prediction window before CA event onset with our model peaking performance at 0.82 one hour before CA.
We also provide a comprehensive analysis of our model through architectural tuning and PaCMAP visualization of patient health trajectory in latent space.
\end{abstract}

% submitted abstract
\begin{comment}
Non-invasive patient monitoring for tracking and predicting adverse acute health events is an emerging area of research. We pursue in-hospital cardiac arrest (IHCA) prediction using only single-channel finger photoplethysmography (PPG) signals. Our proposed two-stage model Feature Extractor-Aggregator Network (FEAN) leverages powerful representations from pre-trained PPG foundation models (PPG-GPT of size up to 1 Billion) stacked with sequential classification models. We propose two FEAN variants (``1H'', ``FH'') which use the latest one-hour and (max) 24-hour history to make decisions respectively. Our study is the first to present IHCA prediction results in ICU patients using only unimodal (continuous PPG signal) waveform deep representations. With our best model, we obtain an average of 0.79 AUROC over 24~h prediction window before CA event onset with our model peaking performance at 0.82 one hour before CA. We also provide a comprehensive analysis of our model through architectural tuning and PaCMAP visualization of patient health trajectory in latent space.
\end{comment}

%%%%%%%%%%%%%%%%%%%%%%%%%%%%%%%%%%%%%%%%%%%%%%%%%%%%%%%%%%%%%%%%%%%%%%%%%%%%%%%%

\section{Introduction}
% wearables, PPG from various physical sites of body, ...
Wearable Health Devices (WHD) are now ubiquitous and offer unique health insights through a combination of an advanced plethora of sensors and embedded AI algorithms.
% , thanks to the plethora of sensors and Artificial Intelligence (AI) algorithms they are equipped with.
Sensors like Photoplethysmography (PPG) are an excellent choice for monitoring cardiac health, specifically, volumetric blood flow in a non-invasive, accessible, and cost-effective manner.
Additionally, PPG technology offers flexibility in terms of anatomical placement of its sensor such as on finger tip~\cite{elgendi2012analysis}, wrist~\cite{moscato2022wrist}, and arm~\cite{zhang2017highly}.
% , finger base (like in smart rings)~\cite{fiore2024use}
% what all physical cardiac parameters can PPG predict
This simple sensor can estimate a wide range of vital signs, including heart rate and respiratory rate~\cite{chen2024adapting,pillai2024papagei,abbaspourazad2023large}, detect underlying health conditions such as Atrial Fibrillation (AF)~\cite{chen2021signal} and Pulseless Electrical Activity (PEA)~\cite{khalili2024detecting}, assess cardiac anatomical parameters such as aortic stiffness~\cite{hellqvist2024estimation} and even unique biometric information~\cite{sancho2018biometric}.
% age~\cite{mathieu2024advanced},
% applications
% such as smartwatches
In addition to wearables, PPG finds application in fatigue detection in driver status monitoring systems~\cite{arakawa2021review} and remote stroke rehabilitation care~\cite{yan2024rehabilitation}.

\begin{figure}
    \centering
    \includegraphics[width=0.85\linewidth]{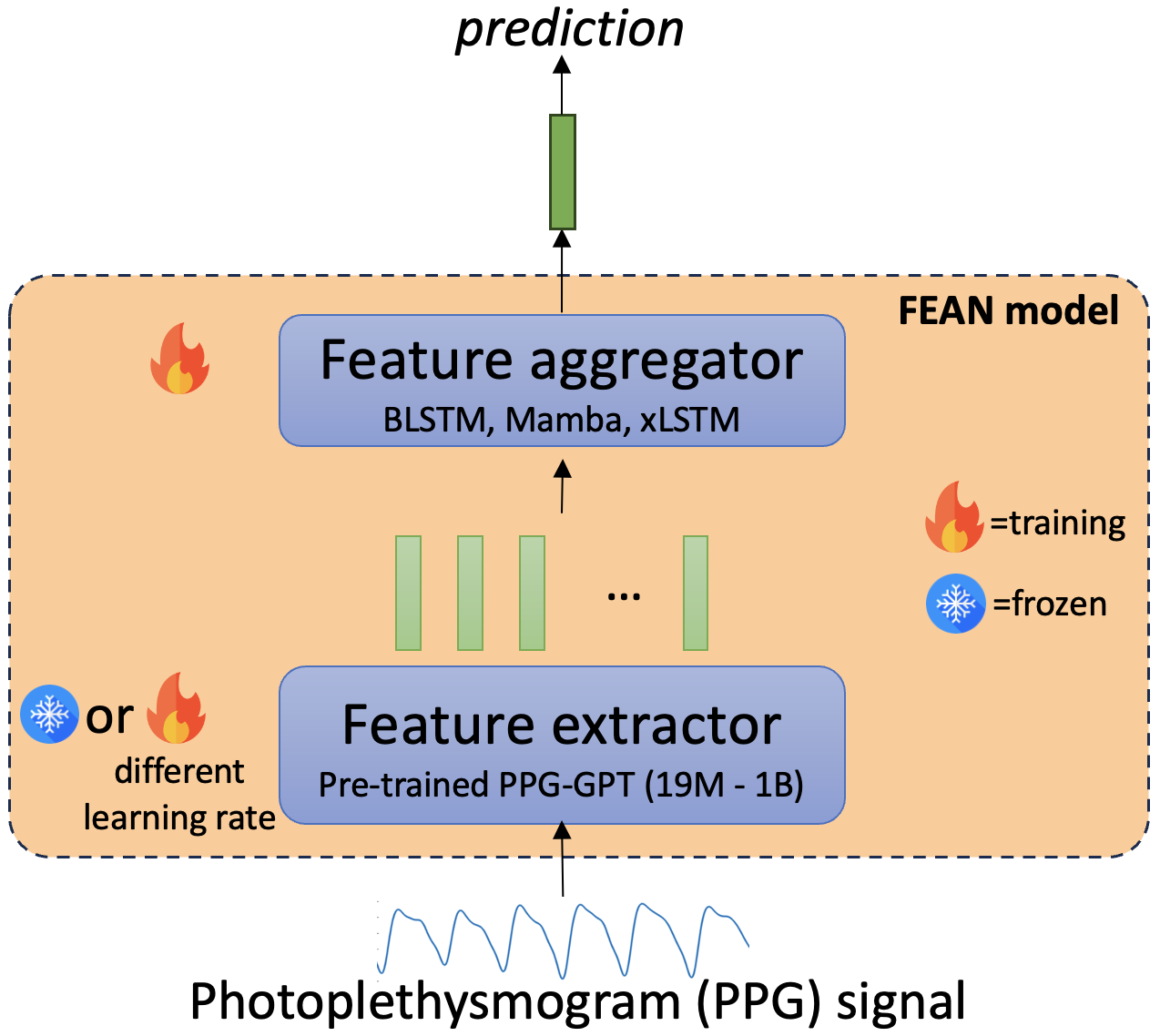}
    \caption{Illustation of Feature Extractor-Aggregator Network (FEAN).
    A pre-trained foundation model extracts features and a sequence model aggregates the embedding sequence into a single vector which finally predicts the target.}
    \label{fig:fean}
\end{figure}

%% CA prediction, critical event prediction
In critical care, PPG plays an important role in predicting acute critical events such as ``code blue''~\cite{bai2016sequence} (hospital calls for emergency resuscitation team~\cite{do2019usefulness}), cardiac arrest~\cite{yan4995228preliminary}, Sepsis~\cite{lauritsen2020early}, stroke~\cite{yan2024rehabilitation}, myocardial infarction~\cite{neha2023automated}.
% bowel disease~\cite{hirten2025physiological}
PPG in combination with other sensor inputs from bedside monitors typically produces \emph{alarms} for possible intervention.
However, this causes issues such as \emph{alarm fatigue}~\cite{hu2019algorithm,drew2014insights}.
To reduce alarm frequency and false positives, combined metrics such as \emph{superalarms}~\cite{hu2012predictive} are devised.
Risk scores like Electronic Cardiac Arrest Risk Triage (eCART)~\cite{bartkowiak2019validating} and Modified Early Warning Score (MEWS)~\cite{tan2022modified} are also used, which consume more multimodal information including vital signs, waveform data (Electrocardiography (ECG), PPG), lab reports, and Electronic Health Record (EHR).
Proactive intervention to treat the underlying stressor before cardiac arrest can help increase monitoring frequency and start active interventions to treat the underlying stress which may reduce mortality rates~\cite{brooks2022optimizing}.

\begin{figure*}
    \centering
    \includegraphics[width=0.9\linewidth]{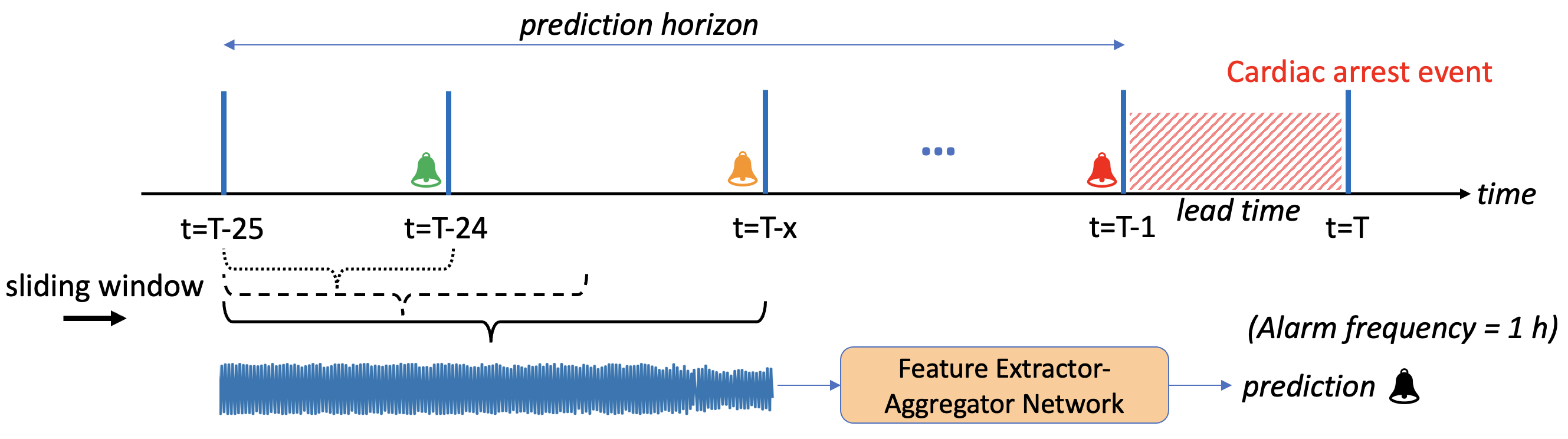}
    \caption{
    Illustration of the overall pipeline (for a case patient).
    1-hour model (``1H'') and Full-history (``FH'') model are trained on 1 h and all history (limited to 24 h max) before a chosen \emph{anchor} timepoint in $[T-25, T-1]$ resp.
    During the evaluation, predictions are made every one hour (alarm frequency) i.e. at T-24, ..., T-1 where t=T is the time of event onset time.
    % Alarm symbols are when alarms may be generated with threshold informed by clinicians.
    }
    \label{fig:pipeline}
\end{figure*}

Our main application for this paper, \emph{automatic IHCA prediction}, has traditionally been approached as a multi-modal or multi-channel problem in previous works.
Kwon et al.~\cite{kwon2020artificial} used 12-lead ECG to predict IHCA and obtained more than 0.9 Area Under the  Receiver Operator Curve (AUROC).
However, Lee et al.~\cite{lee2023real} used single-lead ECG (more feasible for real-time monitoring) and derived Heart Rate Variability (HRV) metrics to obtain 0.88 AUROC.
Another study~\cite{yijing2022prediction} used various invasive and non-invasive vital signs like blood pressure to develop Cardiac Arrest Prediction Index (CAPI) and obtain more than 0.9 AUROC.
Lee et al.~\cite{lee2024prediction} trained a multimodal model using demographics and vital signs (processed by Long-Short Term Memory (LSTM) Deep Neural Network (DNN)) and achieved AUROC of 0.80.
Yan et al.~\cite{yan4995228preliminary} is another multimodal model that additionally uses precursor event (e.g., respiratory failure) onset information.
In Edgar et al.~\cite{inducedCA2024}, authors present preliminary results on detecting induced Out-of-Hospital Cardiac Arrest (OHCA) on short segments of continuous waveform data.

% % common PPG features
Powerful representations of PPG signals are indispensable for achieving state-of-the-art (SOTA) performance on IHCA prediction.
Conventional features of PPG including the second derivative wave of fingertip photoplethysmography (SDPTG)~\cite{takazawa1998assessment}, morphological features~\cite{finnegan2023features}, and short-time Fourier transform~\cite{chen2021signal} (STFT) have been used in the past.
PPG foundation models (FM)~\cite{hu2024foundation} (for e.g., PPG-GPT~\cite{chen2024adapting}, PaPaGei~\cite{pillai2024papagei},) are an emerging choice of feature extractors.
PaPaGei is trained on more than 50,000 hours of data to predict signal quality and distinguish between morphological features of different signals, while the PPG-GPT model~\cite{chen2024adapting} is trained on 2.2 million hours and is the largest available in the literature.
% more FMs
Another model called SiamQuality~\cite{Ding_2024} uses contrastive learning and leverages low and high-quality PPG signals using available quality labels.
A work following Apple Heart and Movement Study (AHMS)~\cite{abbaspourazad2023large} proposed a joint encoder trained with PPG and ECG.

\begin{figure*}[htbp]
    \centering
    % First Subfigure
    \begin{subfigure}[b]{0.92\columnwidth}
        \centering
        \includegraphics[width=\textwidth]{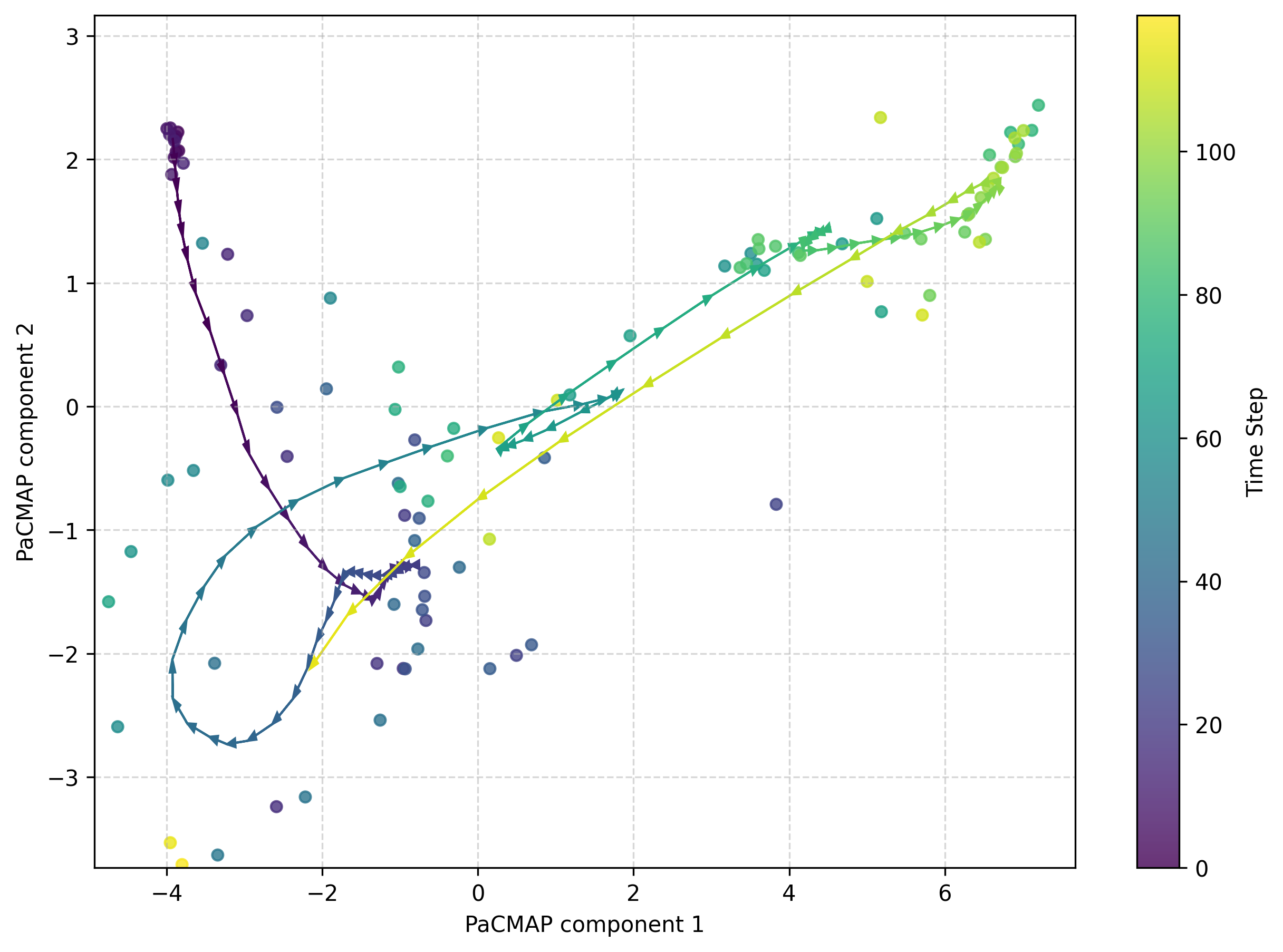}
        \caption{Patient 1}
        \label{fig:patient1}
    \end{subfigure}
    \hfill % Adds horizontal space between the subfigures
    % Second Subfigure
    \begin{subfigure}[b]{0.92\columnwidth}
        \centering
        \includegraphics[width=\textwidth]{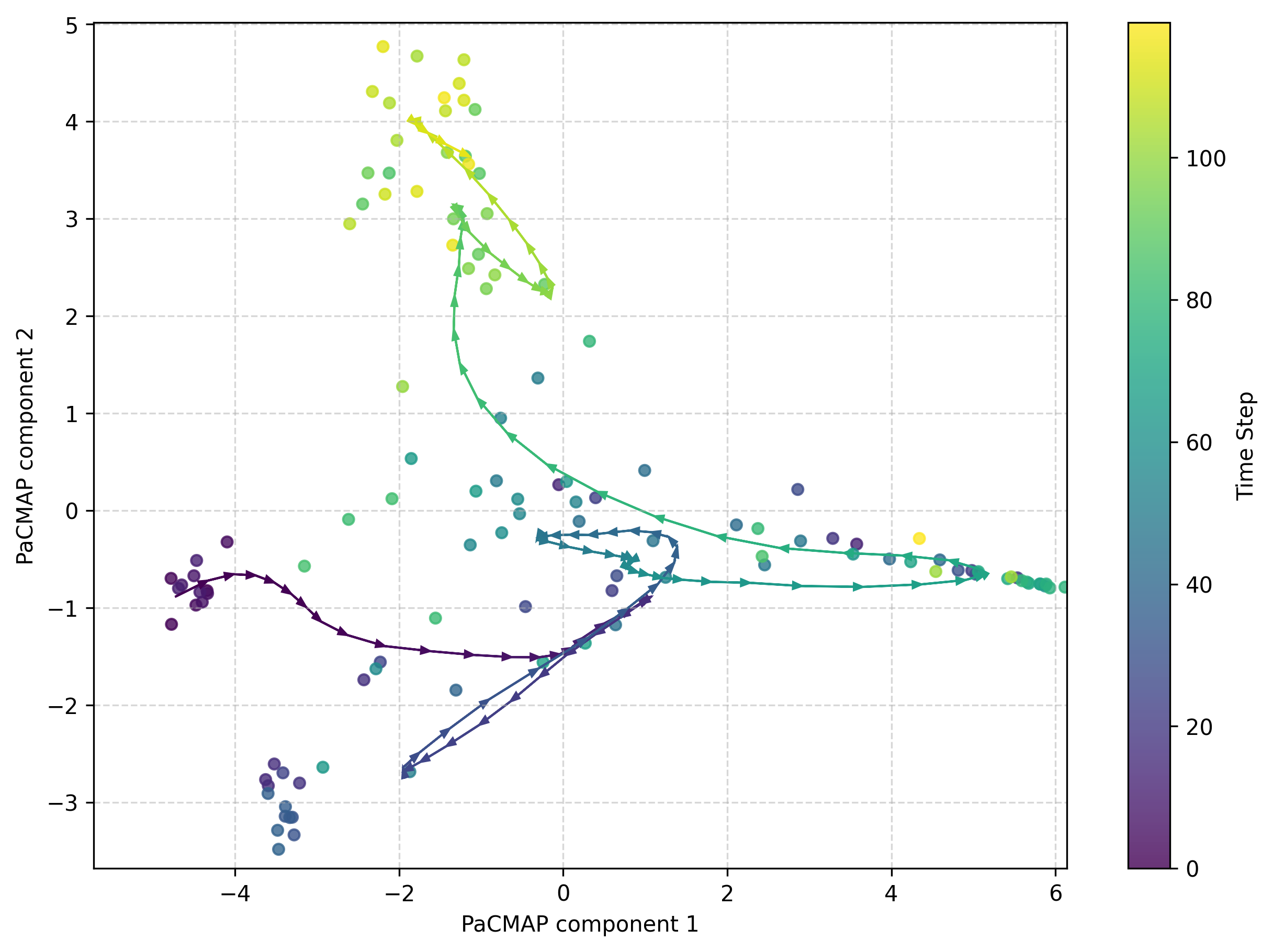}
        \caption{Patient 2}
        \label{fig:patient2}
    \end{subfigure}
    \caption{Illustration of moving average of PaCMAP mapping of PPG-GPT embeddings of 1 hour PPG data before cardiac arrest event. Each time point corresponds to a 30-second segment. Note that the trajectory follows a defined path, potentially capturing the patient's health trajectory in the latent space.}
    \label{fig:trajectory}
\end{figure*}

% weak para? ->contribution summarization to highlight technical design; shout out the exp results.
Our study is the first to present results on the prediction of cardiac arrest (CA) in ICU patients using solely continuous PPG waveform data.
Contrary to the prior work in \cite{lee2024prediction}, our models can ingest up to 24 hours of patient history.
In addition, we demonstrate the utility of PPG foundation model representations up to 1 billion size model - the largest PPG model utilized for downstream tasks.
In the next section, we describe the methodology.

\section{Methodology}

\subsection{Feature Extractor-Aggregator Network (FEAN)}
\label{sec:fean}
For automatic IHCA prediction, we propose to use a simple two-stage cascade model called \emph{Feature Extractor-Aggregator Network} (FEAN) (See Fig.~\ref{fig:fean}).
In the first stage of this model, we utilize a pre-trained representation learning model - in our case, the PPG-GPT generative foundation model.
This \emph{feature extractor} may be frozen or tuned with moderate fine-tuning hyper-parameters.
In the second stage, the \emph{feature aggregator} uses a sequence-to-sequence DNN and combines/pools the \emph{sequential output} from the previous stage into a single vector.
We choose a separate sequential model for the second stage instead of \emph{static} pooling methods~\cite{kim2024universal} since we aim to model sequential dynamics in the output of the first stage.
The final obtained embedding is followed by the binary cross-entropy objective.

\subsubsection{Feature extractor: PPG-GPT}
\label{sec:ppggpt}
The PPG-GPT model was introduced in \cite{chen2024adapting}.
It is an attention-based~\cite{vaswani2017attention} generative model trained to predict the next patch of the input signal that corresponds to one second (or 40 samples for a sampling frequency of 40~Hz).
It was pre-trained on 200M 30~s segments from UCSF data~\cite{drew2014insights,Ding_2024}.
The model sizes available are 19M, 85M, 345M, and 1B.
It is a highly efficient model that uses flash attention~\cite{dao2022flashattention} and is performant based on results on various downstream tasks including false arrhythmia alarm detection.
For feature extraction, we use the last timestamp from the hidden layer output of the last layer.
For input signals longer than 30~s, we divide signals into 30~s chunks and process them independently.

\subsubsection{Feature aggregator}
For aggregating sequential features obtained from the feature extractor model, we experiment with three models: (1) Bi-directional LSTM (BLSTM) with attention (BLSTM-Att)~\cite{schmidhuber1997long}, (2) Mamba~\cite{gu2023mamba}, and (3) xLSTM~\cite{beck2024xlstm}.
The model definitions are available at this repository~\footnote{\url{https://github.com/saurabh-kataria/CA-PPG-FM}}.
The attention in BLSTM-Att is a simple learned weighted sum followed by softmax operation.
For most experiments, we chose BLSTM-Att since we found it to be the most reliable and versatile for our low-resource prediction problem.
Mamba~\cite{gu2023mamba} is a selective state space model (SSM).
Unlike traditional sequence models, Mamba dynamically adjusts its state transitions based on input content through hardware-optimized parallel scanning, enabling linear-time complexity for long sequences.
xLSTM~\cite{beck2024xlstm} extends traditional LSTM architecture with matrix-based memory and exponential gating mechanisms.
The architecture is equipped with a novel gating mechanism (sigmoid for retention, exponential for promotion).
For all feature aggregators, we choose architectural hyper-parameters close to default values obtained from individual repositories with the additional constraint that the number of trainable parameters is approximately 10~M.

\subsection{Data description}
For training and testing, we use the University of California San Francisco (UCSF) Intensive Care Unit (ICU) dataset~\cite{drew2014insights,Ding_2024}.
Similar to the Medical Information Mart for Intensive Care (MIMIC) database~\cite{johnson2023mimic}, it consists of various types of alarms from bedside patient monitors, EHR variables, and continuous physiological signals (invasive, non-invasive) measured in the ICU settings.
This dataset is suitable for In-hospital Cardiac Arrest (IHCA) prediction as it captures data from bedside monitors and includes the signal of interest: finger PPG (the most common form factor in clinical settings~\cite{khalili2024detecting}), and rigorous clinical adjudication of cardiac arrest events.
Note that modeling other modalities might improve performance, but we restrict our approach to using PPG only to minimize reliance on additional inputs and enhance the versatility of the developed technology.
We work with a data subset comprising 200 positive/case patients and 1000 negative/control patients, thereby resulting in a 1:5 imbalance ratio.
The (adult) patients' data has a median age of 63.0 and a standard deviation of 16.3.
Furthermore, similar to \cite{lee2023real}, we retain only 24-hour data up to one hour before cardiac arrest (i.e. T-25 to T-1) for case patients and the last 24-hour ICU data for control patients.
For time gaps when the signal is absent or flat, we simply insert zeroes.

\subsection{1-hour and full-history (1H, FH) FEAN variants}
Fig.~\ref{fig:pipeline} illustrates our pipeline for training and evaluation, which shares similarities to the experimental setup of prior works~\cite{yijing2022prediction}.
The figure focuses on case patients for which t=T is the time of cardiac arrest.
For practicality, our \emph{lead time} is 1 hour.
% which gives at least 1 hour for clinical intervention.
The prediction horizon is 24 hours i.e. we train our model to predict CA within the next 24 hours.
Our training and evaluation scheme follows this.
We propose two variants of our FEAN framework: the 1H model and the FH model.
1H model (or 1-hour model) is a fixed duration model that uses the latest 1 hour of data of a patient to make a prediction.
During training, a random 1-hour subsequence is sampled from $[T-25, T-1]$.
FH model (or full history model) is a variable-length model that uses all the history of patients available to make the prediction (limited to the past 24 hours).
During training, an anchor point is selected from $[T-25, T-1],$, and all history is used.
During the evaluation of both models, alarm frequency is set to 1 hour i.e. every 1 hour a prediction is made.
As stated earlier, 1H uses the latest 1 hour data while FH uses all history.
To simplify the reporting of test results, we evaluate at $t=T-i$ where $i\in 24...1$ since we access to ground truth.

\begin{table}
\centering
\caption{
Results for 1H model using BLSTM-Att feature aggregator and comparison with baseline.
% 1H model uses the latest 1-hour signal available for prediction.
% Results using BLSTM-Att model and comparison with baseline (1H model denotes the prediction horizon set to be 1 hour.)
}
\label{tab:baseline}
\begin{tabular}{|l|c|c|c|c|c|c|}
\hline
& \multicolumn{2}{|c|}{1H model}\\
\hline
Features & AUROC & AUPRC  \\
\hline
Random & 0.500 & 0.106   \\
\hline
Morphological features & 0.553 & 0.146  \\
\hline
STFT & 0.610 &  0.179   \\

\hline
\hline
GPT-19M (frozen) & 0.662 & 0.206  \\
\hline
GPT-85M (frozen) & 0.653 &  0.185   \\
\hline
GPT-345M (frozen) & 0.642 &  0.142 \\
\hline
GPT-1B (frozen) & 0.654 & 0.177    \\

\hline
\hline
GPT-19M & 0.735 & 0.269 \\
\hline
GPT-85M & 0.786 &  0.329 \\
\hline
GPT-345M & \textbf{0.791} &  \textbf{0.338}   \\
\hline
GPT-1B & 0.749 & 0.272   \\
\hline
\end{tabular}
\end{table}

\section{Results}

\begin{figure*}
    \centering
    \includegraphics[width=0.62\linewidth]{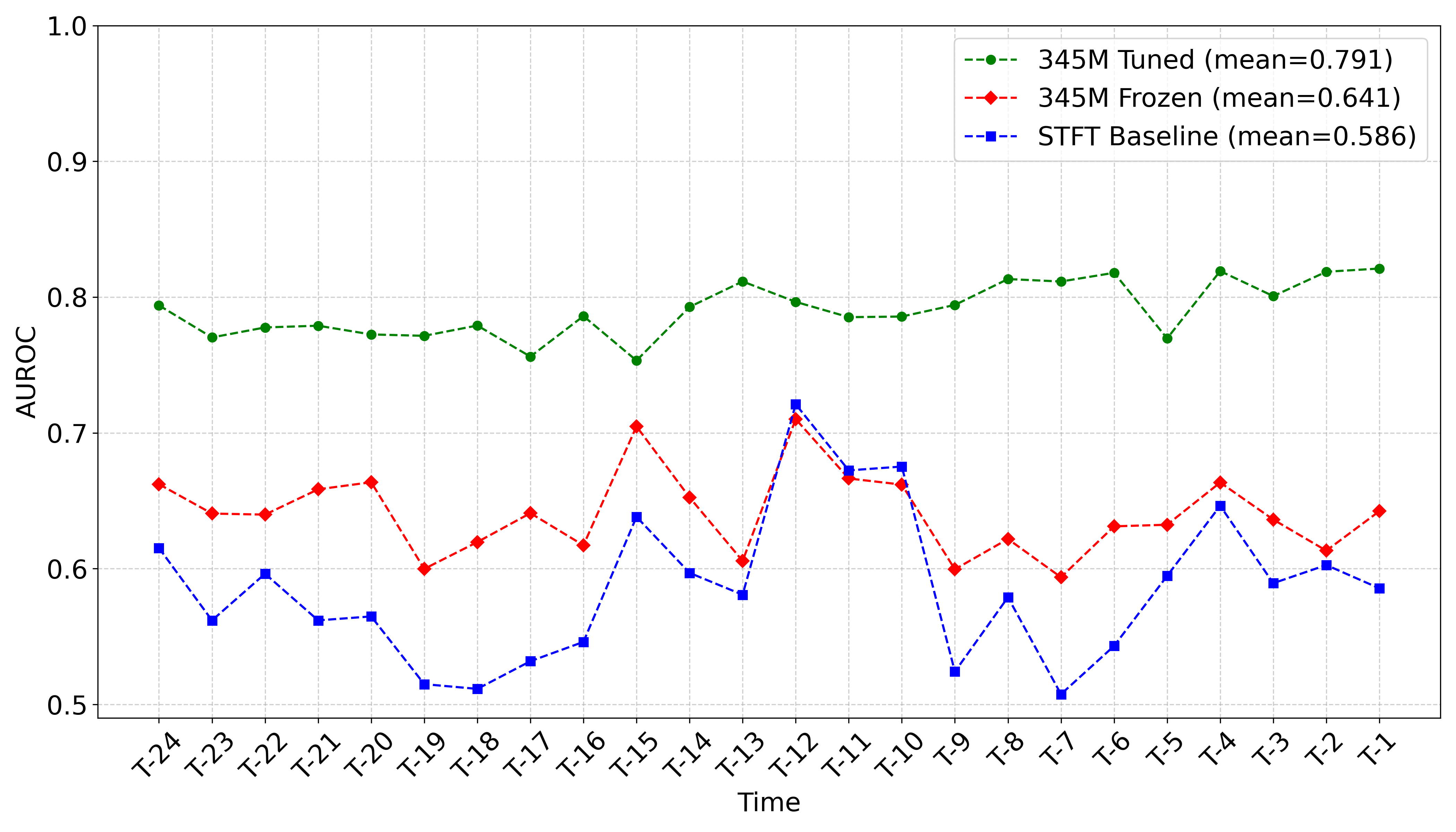}
    \caption{Hourly AUROC for STFT baseline and 1H 345M (both feature extractor tuned and frozen).
    For the feature extractor-tuned model, overall performance is best (printed in legend) and improves when getting closer to the event onset time.
    }
    \label{fig:plot}
\end{figure*}

\subsection{Visualization of embeddings}
We first demonstrate that the deep representations obtained from PPG-GPT for long signals capture temporal dynamics, even though the signal is processed with 30~s independent chunks.
In Fig.~\ref{fig:trajectory}, we illustrate the moving average of 2-D embeddings of 1-hour PPG data before cardiac arrest for two patients. To obtain low-dimensional representations, we use PaCMAP~\cite{wang2021understanding} with default parameters.
Each data point corresponds to a 30~s non-overlapping segment.
% For smoothing the trajectory, we employ the Savitzky-Golay (SG) filter.
% ~\cite{savitzky1964smoothing}
In the moving average visualization, we apply the Savitzky-Golay filter to the embeddings to retain local trends while suppressing high-frequency noise inherent in model outputs.
The trajectory reveals that as time progresses (color scale lightens), embeddings follow a defined path, potentially capturing the patient’s health deterioration in latent space.
While positional embeddings in GPT trivially encode time within 30~s chunks, our results suggest they implicitly capture inter-chunk temporal dependencies.

\subsection{Comparison of 1H model and baseline}
In Table~\ref{tab:baseline}, we detail the results of the 1H model and compare them with three baselines: random features, PPG morphological features (using pyPPG library~\cite{goda2024pyppg}), and STFT features.
Similar to the parallel work of \cite{yan4995228preliminary}, our study also explored morphological features of PPG as important features.
For training the feature aggregator, we use 100 epochs with a batch size of 64, a learning rate of 0.0002, a weight decay of 0.001, and the MARS~\cite{yuan2024mars} optimizer.
Tuning the pre-trained PPG-GPT model along with the feature aggregator is challenging.
Therefore, when PPG-GPT is fine-tuned in an experiment, it follows moderate hyper-parameters such as a low learning rate of 0.00001 and a high weight decay of 0.02.
The obtained area under the ROC curve and precision-recall curve i.e. AUROC and AUPRC are tabulated.
By tuning the foundation model, we obtain better results.
Also, the performance correlates with the size of the foundation model.
These metrics are averaged over the 24-hour prediction window and thus, we report the average of 24 values.
The raw values are tabulated in the Appendix Table~\ref{tab:raw} and the ROC curve for the best model is in Appendix Fig.~\ref{fig:roc}.
For further insights, we plot the values in Fig.~\ref{fig:plot}.
Our curves are similar to ones obtained in the prior work~\cite{lee2024prediction}.

\begin{table}
\centering
\caption{Exploring different architectures for feature extractor using \emph{tuned} 19M and 1B GPT feature extractor.}
\label{tab:arch}
\begin{tabular}{|l|c|c|c|c|}
\hline
& \multicolumn{2}{|c|}{1H model} \\
\hline
 & AUROC & AUPRC  \\
\hline
\multicolumn{3}{|l|}{\textit{GPT-19M FE}} \\
\hline
BLSTM & 0.681 &  0.224   \\
\hline
BLSTM-Att & \textbf{0.735} &  \textbf{0.269}   \\
\hline
Mamba & 0.722 &  0.252   \\
\hline
xLSTM & 0.670 &  0.227   \\
\hline
\hline
\multicolumn{3}{|l|}{\textit{GPT-1B FE}} \\
\hline
BLSTM & 0.693 &   0.202  \\
\hline
BLSTM-Att & \textbf{0.749} &  \textbf{0.272} \\
\hline
Mamba & 0.696 & 0.218  \\
\hline
xLSTM & 0.652 &  0.1845 \\
\hline
\end{tabular}
\end{table}

\begin{table}
\centering
\caption{Results for FH model where feature extractor PPG-GPT is kept frozen (due to computational challenges) and feature aggregator is BLSTM-Att.}
\label{tab:full}
\begin{tabular}{|l|c|c|}
\hline
& \multicolumn{2}{c|}{FH model} \\
\hline
  &  AUROC & AUPRC  \\
\hline
(1H Baseline) GPT-19M + BLSTM-Att & 0.662 & 0.206  \\
\hline
FEAN model (FE frozen)  &   &   \\
\hline
GPT-19M + BLSTM-Att & 0.763 &  0.340 \\
\hline
GPT-85M + BLSTM-Att & 0.738 & 0.333  \\
\hline
GPT-345M + BLSTM-Att & 0.762 &  0.353 \\
\hline
GPT-1B + BLSTM-Att & \textbf{0.764} &  \textbf{0.354}   \\
\hline
\end{tabular}
\end{table}

\subsection{Comparison of different architectures}
In Table~\ref{tab:arch}, using two feature extractors small GPT-19M and large GPT-1B, we evaluate different feature aggregators: BLSTM, BLSTM with attention (BLSTM-Att) (Sec.~\ref{sec:fean}), Mamba, and xLSTM.
Due to low-resource data availability, we notice a high variance of our model performance w.r.t. the choice of hyperparameters (data design as well as training methodology).
Mamba shows promising results and we defer experimenting with its architecture in future work.

\subsection{Full-history}
In Table~\ref{tab:full}, we explore whether the full history model can deliver better performance than the 1-hour model.
Due to the computational challenges (GPU memory) of processing 24-hour-long signals and fine-tuning the foundation model, we report the results of this experiment on the frozen scenario only.
We observe much better performance compared to the 1H frozen baseline.
The overall best performance is obtained by the 1B feature extractor.

\section{Conclusion}
% need to make this longer
In this work, we addressed the prediction of In-Hospital Cardiac Arrest using only non-invasive PPG signals and obtained an average AUROC of 0.79 (comparable to prior works).
% generalize to OHCA
Furthermore, we provide a proof-of-concept for using unimodal biosignal foundation models to match the performance of simpler multi-modal systems in prior work.
To fully leverage the scale of PPG FMs, we separately trained a 1B version of PPG-GPT for this work which paves the way for scaling law studies.

One limitation of IHCA works including ours is the number of positive case patients (200).
In the future, we can leverage MIMIC databases~\cite{johnson2023mimic} or utilize simulation of abnormal cardiac states~\cite{khalili2024detecting}.
We can also pursue survival models that take into account the time of event onset (time-to-event modeling~\cite{kim2020development,salas2015cumulative}) as well or explore multi-resolution features like in \cite{kim2023explainable} - gathering statistics of features at different periods or time-frequency features~\cite{xiao2020generalizability}.
Another direction would be to compare performance with other foundation models like PaPaGei~\cite{pillai2024papagei}.
% We could also pursue conditioning with auxiliary information or joint modeling to handle signal quality compensation and domain adaptation.
% ~\cite{kataria2023self}
% ~\cite{kataria2022joint}
Finally, we can conduct cross-institutional testing~\cite{xiao2020generalizability} to test generalization to other bedside capture devices and patient demographics.

% \section{Conclusion}
% x

% \clearpage

\section*{APPENDIX}

\begin{table}[h!]
\centering
\caption{Raw AUROC values obtained for three models (STFT Baseline, 345M Frozen, 345M Tuned).
The lack of consistency in patterns is 
}
\label{tab:raw}
\begin{tabular}{|c|c|c|c|}
\hline
Time of prediction  &  STFT & 345M-Frozen & 345M-Tuned  \\
\hline
T-24 & 0.615 & 0.662 & 0.794 \\
\hline
T-23 & 0.562 & 0.641 & 0.770 \\
\hline
T-22 & 0.596 & 0.640 & 0.778 \\
\hline
T-21 & 0.562 & 0.658 & 0.779 \\
\hline
T-20 & 0.565 & 0.664 & 0.772 \\
\hline
T-19 & 0.515 & 0.600 & 0.771 \\
\hline
T-18 & 0.511 & 0.620 & 0.779 \\
\hline
T-17 & 0.532 & 0.641 & 0.756 \\
\hline
T-16 & 0.546 & 0.617 & 0.786 \\
\hline
T-15 & 0.638 & 0.705 & 0.753 \\
\hline
T-14 & 0.597 & 0.652 & 0.793 \\
\hline
T-13 & 0.581 & 0.606 & 0.812 \\
\hline
T-12 & \textbf{0.721} & \textbf{0.710} & 0.796 \\
\hline
T-11 & 0.672 & 0.666 & 0.785 \\
\hline
T-10 & 0.675 & 0.662 & 0.786 \\
\hline
T-9 & 0.524 & 0.600 & 0.794 \\
\hline
T-8 & 0.579 & 0.622 & 0.813 \\
\hline
T-7 & 0.508 & 0.594 & 0.811 \\
\hline
T-6 & 0.543 & 0.631 & 0.818 \\
\hline
T-5 & 0.595 & 0.632 & 0.770 \\
\hline
T-4 & 0.646 & 0.663 & 0.819 \\
\hline
T-3 & 0.589 & 0.636 & 0.801 \\
\hline
T-2 & 0.603 & 0.613 & 0.819 \\
\hline
T-1 & 0.586 & 0.642 & \textbf{0.821} \\
\hline
\hline
Mean (All) & 0.586 & 0.641 & \textbf{0.791} \\
\hline
\hline
Mean (Last 12h) & 0.603 & 0.639 & \textbf{0.803} \\
\hline
\hline
Mean (Last 6h) & 0.594 & 0.636 & \textbf{0.808} \\
\hline
\end{tabular}
\end{table}

\begin{figure}[h!]
    \centering
    \includegraphics[width=0.9\linewidth]{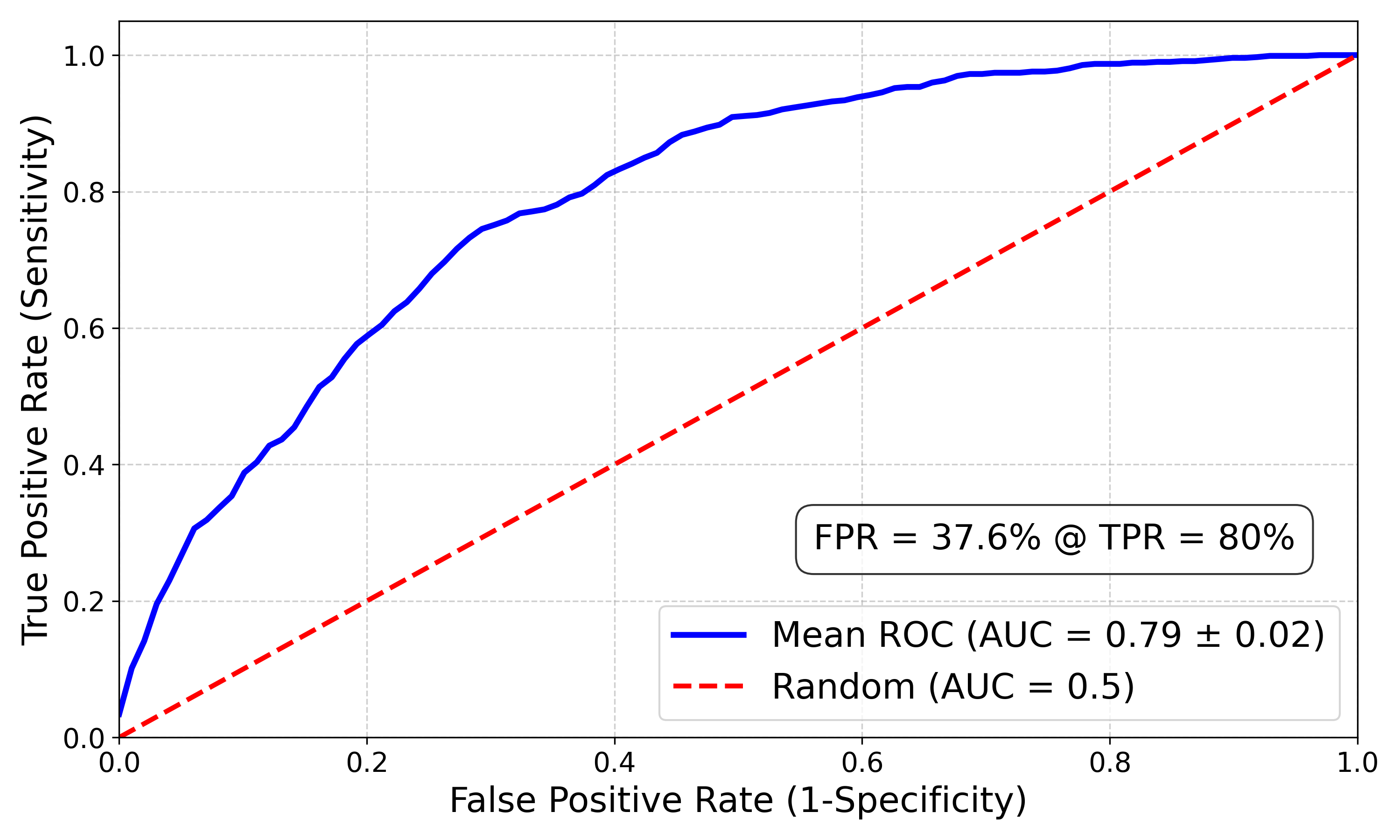}
    \caption{Receiver Operating Curve (ROC) values averaged over 24-hour prediction window.
    Here, prevalence is 11\%.
    The optimal threshold will depend on metrics to minimize, for instance, false alarms to reduce alarm fatigue~\cite{hu2012predictive}.
    }
    \label{fig:roc}
\end{figure}

% \section*{ACKNOWLEDGMENT}
% Put sponsor acknowledgments in the unnumbered footnote on the first page.

% \clearpage

\bibliographystyle{unsrt}      % You can choose a different style if you prefer
\bibliography{root}      % Ensure 'references' matches your .bib filename without the extension

\end{document}